\documentclass[10pt, a4paper]{article}

\usepackage[final]{lrec-coling2024} 

\usepackage{amsfonts}
\usepackage{bbm}
\usepackage{color}
\usepackage{times}
\usepackage{soul}
\usepackage{url}
\usepackage[utf8]{inputenc}
\usepackage{graphicx}
\usepackage{amsmath}
\usepackage{amsthm}
\usepackage{booktabs}
\usepackage{algorithm}
\usepackage{algorithmic}
\usepackage{amsthm,amssymb,mathrsfs}
\usepackage{multirow}
\usepackage{stfloats}
\usepackage{bm}
\usepackage{changes}
\usepackage{makecell}
\usepackage{balance}
\usepackage{tcolorbox}
\usepackage{xcolor}

\title{Biomedical Entity Linking as Multiple Choice Question Answering}

\name{Zhenxi Lin$^1$, Ziheng Zhang$^1$, Xian Wu, Yefeng Zheng} 

\address{
Jarvis Research Center, Tencent YouTu Lab, Shenzhen, China\\
}

\abstract{
Although biomedical entity linking (BioEL) has made significant progress with pre-trained language models, challenges still exist for fine-grained and long-tailed entities. 
To address these challenges, we present BioELQA, a novel model that treats \underline{Bio}medical \underline{E}ntity \underline{L}inking as Multiple Choice \underline{Q}uestion \underline{A}nswering.
BioELQA first obtains candidate entities with a fast retriever, jointly presents the mention and candidate entities to a generator, and then outputs the predicted symbol associated with its chosen entity.
This formulation enables explicit comparison of different candidate entities, thus capturing fine-grained interactions between mentions and entities, as well as among entities themselves.
To improve generalization for long-tailed entities, we retrieve similar labeled training instances as clues and concatenate the input with retrieved instances for the generator.
Extensive experimental results show that BioELQA outperforms state-of-the-art baselines on several datasets.
\\ \newline }

\begin{document}

\maketitleabstract

\footnotetext[1]{Correspondence to \texttt{chalerislin@tencent.com} and \texttt{zihengzhang@tencent.com}.}

\section{Introduction}
\label{sec:intro}

Biomedical entity linking (BioEL) refers to mapping biomedical mentions to standard entities in an ontology, such as the Unified Medical Language System (UMLS)~\cite{bodenreider2004unified}, which is essential for various downstream tasks, including automatic diagnosis~\cite{yuan2021efficient,shi2022understand}, drug-drug interaction prediction~\cite{li2023drug,zhang2023emerging}, and knowledge graph alignment~\cite{xiang2021ontoea,lin2022multi}.
Unlike entity linking in the general domain, biomedical entities often have various names, including synonyms and morphological variations, such as "motrin" is also referred to as "ibuprofen". Additionally, similar surface forms of biomedical entities can have distinct meanings, such as "Type 1 Diabetes" and "Type 2 Diabetes".

Existing methods can be mainly categorized into two types.
One is the discriminative methods~\cite{sung2020biomedical, lai2021bert, liu2021self, wu2023modeling}, which employed BERT-based models to encode mentions and entities into the same embedding space and disambiguated mentions by nearest neighbors, or further applied a cross-encoder to rerank top candidates by capturing fine-grained mention-entity interactions~\cite{angell2021clustering, zhu2021enhancing, xu2023improving}.
Another one is the generative methods~\cite{yan2020knowledge,yuan2022biobart,yuan2022generative} that directly generated linked entities using text-to-text pre-trained language models, such as BART~\cite{lewis2020bart}, thereby circumventing the need for negative sample mining.


However, BioEL remains challenging due to the fine-grained and long-tailed entities.
First, previous methods focused on mention-entity interactions, but generally neglected fine-grained interactions between candidate entities (\textit{i.e.}, entity-entity interactions) and struggled with ambiguous mentions of multiple candidates with similar surface forms~\cite{xu2023improving}.
For example, when the mention ``haemoglobin'' is compared to two closely related candidate entities, ``haemoglobin c'' and ``hemoglobin'', the high lexical similarity between them confuses the models, leading to potential mismatch as ``haemoglobin c''. 
Incorporating such entity-entity interactions can provide a more holistic and nuanced representation of interactions among entities.
Furthermore, the number of candidate entities can be large, and they often follow long-tailed distribution~\cite{kim2021read}, further exacerbating the difficulty of BioEL.
Previous studies have demonstrated performance improvement for long-tailed entities using auxiliary information, such as entity descriptions and synonyms~\cite{varma2021cross,yuan2022generative}.
However, collecting such data is labor-intensive, which limits its applicability.


\begin{figure*}[!ht]
    \centering
    \includegraphics[width=1.0\linewidth]{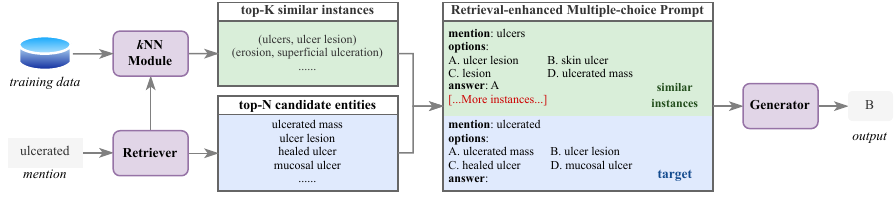}
    \caption{The overview of the proposed BioELQA.}
    \label{fig:model}
\end{figure*}

To tackle the aforementioned challenges, we present a novel model called BioELQA, which aims at capturing the fine-grained interactions between entities and enhancing the generality of long-tailed entities.
We achieve this by reframing BioEL as a Multiple Choice Question Answering (MCQA) task.
Given a mention, we first use a bi-encoder retriever to efficiently retrieve top-$N$ candidate entities from the ontology as the ``answer options''.
Then the mention and its answer options are all fed to a generator as a multiple choice prompt, with each answer associated with a symbol (\textit{e.g.}, "A", "B", "C").
The prompt is structured so that the generator outputs the symbol associated with its chosen answer option.
This framework enables explicit comparison and contrast among different answer options, effectively modeling the mention-entity interactions and entity-entity interactions.
Besides, by directly generating answer symbols instead of entity names, we can avoid generating entities that do not exist in the ontology and reduce reliance on normalization strategies~\cite{robinson2022leveraging}.
Inspired by the recent success of retrieval-augmented model~\cite{khandelwal2020nearest}, we further leverage a $k$NN module to retrieve semantically similar instances from the training data to formulate a retrieval-enhanced prompt, which empowers the model to reference similar contexts as cues for prediction, to improve the robustness and generalization for long-tailed entities.
We experimentally validate the effectiveness and superiority of BioELQA as it achieves state-of-the-art performance on several benchmark datasets.
The source code and curated prompts are publicly accessible at \url{https://github.com/lzxlin/BioELQA}.

\section{Method}
\label{sec:method}

The goal of BioEL is to link a given mention $m$ to its correct entity $e$ in an ontology $\mathcal{E}$.
The overall architecture of the proposed BioELQA is shown in Figure~\ref{fig:model}. 
Firstly, a retriever (\S \ref{ssec:retriever}) selects the top-$N$ candidate entities for mention $m$, and a $k$NN module (\S \ref{ssec:retrieval-module}) provides top-$K$ similar instances from the training corpus as contextual clues.
These components form a retrieval-enhanced multiple-choice prompt, which is then processed by a generator (\S \ref{ssec:generator}) to produce the final answer.

\subsection{Retriever}
\label{ssec:retriever}

Following the previous work~\cite{zhu2021enhancing,xu2023improving}, we employ a bi-encoder based on SapBERT~\cite{liu2021self} to generate dense vectors for both mentions and entities.
The mention embedding $f(m)$ of mention $m$ is formulated as:
\begin{gather}
    f(m) = \texttt{SapBERT}(m)\texttt{[CLS]}\label{eq:emb_m},
\end{gather}
where $\texttt{[CLS]}$ denotes the special token that is used to derive a fixed-size vector.
The entity embedding $f(e)$ of entity $e$ is computed similarly.
The score of a mention-entity pair $(m,e)$ is denoted as follows:
\begin{gather}
    s(m, e) = g(f(m), f(e)),
\end{gather}
where $g$ is the cosine similarity. At inference time, we precompute $f(e)$ for each $e\in\mathcal{E}$ and use Faiss~\cite{johnson2019billion} for fast retrieval.

\noindent\textbf{Training}. We use contrastive learning to train the retriever, which aims at optimizing the agreement between true mention-entity pairs and the disagreement between false ones. The loss for each true pair $(m,e)$ is computed as:
\begin{equation}
\begin{small}
\begin{aligned}
    \mathcal{L}(m,e)=
    -\log
    \left( 
    \frac{\delta(m,e)}
    {\delta(m,e) + \sum\limits_{e'\in{\mathcal{H}(e)}}\delta(m,e')} 
    \right),
    \label{eq:loss}
\end{aligned}
\end{small}
\end{equation}
where $\delta(m,e)=\exp(s(m,e)/\tau)$, $\tau$ is a temperature hyper-parameter, and $\mathcal{H}(e)\subset\mathcal{E}\backslash\{e\}$ is a set of negatives that excludes $e$. We obtain $\mathcal{H}(e)$ by combining in-batch negative sampling and hard negative sampling (\textit{i.e.}, highest-scoring incorrect entities), which has been shown beneficial for entity retrieval~\cite{wu2020scalable, gao2021simcse}.

\subsection{Generator}
\label{ssec:generator}

\noindent\textbf{Multiple-Choice Prompting (MCP)}.
Given $m$ and top-$N$ candidate entities from the retriever, we reformulate BioEL as a multiple-choice question answering task. Specifically, it involves selecting the correct answer from a set of answer options (\textit{i.e.}, candidate entities) based on the question (\textit{i.e.}, mention).
There are two ways for answer generation: 1) the use of closed prompts~\cite{xu2023improving} that fill in the blanks of a sentence through masked language modeling; 2) the use of generative models~\cite{yuan2022biobart,yuan2022generative} that directly generate entity names.
However, the former relies on well-designed prompts and continuous pre-training, and the latter could generate non-existent entities and require complex normalization strategies. An appropriate multiple-choice prompt is to jointly present the question and answer options to the model and guide it to output the symbol (\textit{e.g.}, “A”) associated with its chosen answer option.

Formally, given $m$ as the question and its corresponding answer options $O=\{o_1, o_2, ..., o_N\}$ sorted in descending order of similarity to $m$, we first map $(m, O)$ to a text sequence using a simple template $\mathcal{T}(\cdot)$: ``mention: $m$ options: A. $o_1$ B. $o_2$ C. $o_3$ D. $o_4$ answer:'' (when $N=4$), as shown in Figure~\ref{fig:model}. 
Then the sequence is fed into a generator $\mathcal{M}$, such as T5~\cite{raffel2020exploring}, to produce answer symbol $a$, where $a$ is the symbol associated with the correct answer.
The training objective of the generator is to maximize the likelihood: $p(a|m, O) = p_\mathcal{M}(\mathcal{T}(m, O))$. 
We train the generator using a standard seq2seq objective, \textit{i.e.}, maximizing the output sequence likelihood with teacher forcing~\cite{sutskever2014sequence}.




In MCP, a mention and its symbol-enumerated candidate answers are all passed to the generator as a single prompt, explicitly modeling both mention-entity and entity-entity interactions.
Moreover, MCP enables the generator to only predict a single token (\textit{e.g.}, "A") instead of entity names, avoiding the generation of invalid entities.
The token-associated answer predicted by the model is the final answer.
The probabilities of these symbols therefore serve as a proxy for the probability of each answer.


Previous studies~\cite{robinson2022leveraging, pezeshkpour2023large} have found that current language models are sensitive to the order of answer options that a slight change in the order can alter the model's answer.
To mitigate this problem and improve robustness, we employ a simple yet effective data augmentation strategy during training.
By randomly swapping the order of $O$, we can construct different $(\mathcal{T}(m, O), a)$ training pairs. Theoretically, each $(m, O)$ can generate $N!$ distinct examples.
Preliminary experiments indicate that augmenting each input with one additional swap is sufficient.
This strategy allows the model to learn the association between symbols and answers, rather than solely memorizing answer positions.

\subsection{$k$NN Module}
\label{ssec:retrieval-module}

Inspired by~\citet{lin2023improving}, we introduce a $k$NN module to enhance the model's generalization capabilities for long-tailed entities, enabling it to reference similar instances from the entire training corpus as prediction clues.

Given $m_i$ in the training data, there are $N$ candidate options $O_i$ and a true answer symbol $a_i$ constructed by MCP (\S \ref{ssec:generator}).
We construct a datastore $\mathcal{D}$ by indexing the training data as a list of key-value pairs $\{(f(m_i), (m_i, O_i, a_i))\}$, where the key is the mention embedding computed in Eq.~(\ref{eq:emb_m}) and the value is a training instance for generator.
Given a mention $x$ as input, we compute its mention embedding $f(x)$ via Eq.~(\ref{eq:emb_m}), query the datastore with $f(x)$ to all keys based on cosine similarity, and obtain the top-$K$ most similar instances $\mathcal{N}=\{(m_{i_1}, O_{i_1}, a_{i_1}), ..., (m_{i_K}, O_{i_K}, a_{i_K})\}$.
Note that during training, as $x$ is already indexed, we filter it from the retrieved results to avoid data leakage.
We convert these instances $\mathcal{N}$ into text sequences using $\mathcal{T}(\cdot)$ individually, and then concatenate them together with the input sequence $\mathcal{T}(x, O)$, where $O$ corresponds to the answer options for $x$. This forms a retrieval-enhanced multiple-choice prompt that is fed into the generator to generate the answer, as shown in Figure~\ref{fig:model}.
Now the training objective of the generator becomes as follows:
\begin{equation}
\begin{small}
\begin{aligned}
    p(a|x, O, \mathcal{N}) = p_\mathcal{M}(
    \mathcal{T}(m_{i_1}, O_{i_1})\oplus a_{i_1}
    \oplus \ldots \oplus\\
    \mathcal{T}(m_{i_K}, O_{i_K})\oplus a_{i_K}
    \oplus
    \mathcal{T}(x, O)),
    \label{eq:enhanced-loss}
\end{aligned}
\end{small}
\end{equation}
where $\oplus$ is concatenation operation and $a$ is the answer symbol of $(x, O)$.
This can choose informative instances for each input dynamically and provides the generator with direct evidence about the input and references to make predictions.
Our approach is similar to few-shot learning~\cite{brown2020language}, yet with a key distinction of our focus on supervised learning, where model parameters are fine-tuned from given instances for performance improvement.

\section{Experiments}
\label{sec:exp}

\subsection{Experimental Setup}
\label{ssec:setup}

\noindent\textbf{Datasets and Evaluation}. 
Three BioEL datasets are adopted, including NCBI~\cite{dougan2014ncbi}, BC5CDR~\cite{li2016biocreative} and COMETA~\cite{basaldella2020cometa}, which focus on different entity types, such as diseases and chemicals. The detailed dataset statistics are listed in Table~\ref{tab:dataset}.
We report accuracy for all methods, and the best results are in \textbf{bold} with the second best results \underline{underlined}.

\begin{table}[h]
    \centering
    \footnotesize
    \renewcommand\arraystretch{1.0}
    \begin{tabular}{@{}lrrrr@{}}
        \toprule
        & NCBI & BC5CDR & COMETA  \\
        \midrule
        Target entities $|\mathcal{E}|$ & 14,967 & 268,162 & 350,830 \\
        Train instances & 5,784 & 9,285 & 13,489 \\
        Dev instances & 787 & 9,515 & 2,176 \\
        Test instances & 960 & 9,654 & 4,350 \\
        \bottomrule
    \end{tabular}
    \caption{Dataset Statistics.}
    \label{tab:dataset}
\end{table}


\noindent\textbf{Implementation Details}.
We use \texttt{t5-base}~\cite{raffel2020exploring} as the generator backbone.
The number of training epochs is 20 and the batch size is 16.
The hyper-parameter $\tau$ is 0.01.
The number of options $N$ and the number of similar instances $K$ are 5 and 3, respectively.
We search learning rate among $\{4\times 10^{-5}, 8\times 10^{-5}, 1\times 10^4, 2\times 10^{-4}, 4\times 10^{-4}\}$ for different datasets.
We use the AdamW optimizer to update model parameters.

\noindent\textbf{Baselines}. 
We compare the proposed BioELQA against previous state-of-the-art BioEL methods, which are classified into three categories:
1) \texttt{discriminative-based} methods that utilize bi-encoders or extend their capabilities with cross-encoders to retrieve relevant entities, including BioSyn~\cite{sung2020biomedical}, ResCNN~\cite{lai2021bert}, SapBERT~\cite{liu2021self}, Clustering-based~\cite{angell2021clustering} and Prompt-BioEL~\cite{xu2023improving};
2) \texttt{generative-based} methods that directly generate the linked entities, including GenBioEL~\cite{yuan2022generative} and BioBART~\cite{yuan2022biobart}; 3) \texttt{LLM-based} methods that have demonstrated impressive capabilities across various tasks, especially in the biomedical domain, with simple instructions~\cite{jahan2023comprehensive}, including GPT3.5,\footnote{\url{https://platform.openai.com/docs/models/gpt-3-5}} PaLM-2~\cite{anil2023palm}, Claude-2,\footnote{\url{https://www.anthropic.com/index/claude-2}} and LLaMA-2-7b/13b~\cite{touvron2023llama}. Note that the results of \texttt{LLM-based} methods are all taken from~\citet{jahan2023comprehensive}.


\subsection{Overall Results}
\label{ssec:result}

Table~\ref{tab:overall-perf} shows that BioELQA outperforms previous baselines on all datasets, demonstrating the effectiveness of our proposed method.
Note that the strongest baseline Prompt-BioEL requires additional synonym corpora for continued pre-training, while BioELQA only leverages the existing training data without the need for additional corpora.
We observe that Claude-2 outperforms other \texttt{LLM-based} methods in all datasets, but it still significantly lags behind fully supervised methods.
This suggests that fine-tuning small-scale models remains an effective choice for BioEL, as they can acquire domain-specific knowledge through parameter tuning.

\begin{table}[ht]
    \centering
    \footnotesize
    \renewcommand\arraystretch{1.0}
    \begin{tabular}{l|c|c|c@{}}
        \toprule
        Models & NCBI & BC5CDR & COMETA \\
        \midrule
        BioSyn & 91.1 & -- & 71.3 \\
        ResCNN & 92.4 & -- & 80.1 \\
        SapBERT & 92.3 & -- & 75.1 \\
        Clustering-based & -- & 91.3 & -- \\
        Prompt-BioEL & \underline{92.6} & \underline{93.7} & \underline{83.7} \\
        \midrule
        GenBioEL & 91.9 & 93.3 & 81.4 \\
        BioBART-base & 89.3 & 93.0 & 79.6 \\
        BioBART-large & 89.9 & 93.3 & 81.8 \\
        \midrule
        GPT3.5 & 52.2 & 54.9 & 43.5 \\
        PaLM-2 & 38.4 & 52.1 & 48.8 \\
        Claude-2 & 70.2 & 78.0 & 53.3 \\
        LLaMA-2-13b & 59.2 & 66.5 & 40.7 \\
        \midrule
        BioELQA & \textbf{93.5} & \textbf{94.5} & \textbf{85.2} \\
        \bottomrule
    \end{tabular}
    \caption{Comparison of different BioEL methods on three public datasets.}
    \label{tab:overall-perf}
\end{table}


\subsection{Ablation Study}
To investigate different components in BioELQA, we compare BioELQA variants without data augmentation (w/o data aug.) in \S \ref{ssec:generator} and without $k$NN module (w/o $k$NN module).
We also explore replacing similar instances with randomly selected examples from the training data (random instances) and requiring the generator to predict complete entity names instead of answer symbols (generate names).
We observe that removing any component resulted in varying degrees of performance degradation, thus highlighting the contributions of these components.
Particularly, generating meaningful entity names is shown more challenging than generating answer symbols alone, as it requires the model to learn from the context, leading to potential performance drops.

\begin{table}[ht]
    \centering
    \footnotesize
    \renewcommand\arraystretch{1.0}
    \begin{tabular}{l|c|c|c@{}}
        \toprule
        Model & NCBI & BC5CDR & COMETA \\ 
        \midrule
        BioELQA & \textbf{93.5} & \textbf{94.5} & \textbf{85.2} \\
        \quad w/o data aug. & 93.1 & 94.4 & 83.4 \\
        \quad w/o $k$NN module & 91.3 & 94.1 & 84.3 \\
        \quad random instances & 91.5 & 94.0 & 84.6 \\
        \quad generate names & 66.9 & 89.1 & 78.5 \\
        \bottomrule
    \end{tabular}
    \caption{Ablation study on three datasets.}
    \label{tab:ablation-study}
\end{table}

\subsection{Case Study}
Table~\ref{tab:case-study} shows two long-tailed entities \textit{uneasy} and \textit{misty vision} to illustrate the amendment by $k$NN module.
Without the $k$NN module, the model tends to link mentions to morphologically similar but incorrect entities. 
Nevertheless, incorporating the $k$NN module allows the model to leverage related instances from the training set, enabling more informative decisions by referencing these instances.


\begin{table}[h]
    \centering
    \scriptsize
    \renewcommand\arraystretch{1.0}
    \begin{tabular}{@{}l|l|l@{}}
        \toprule
        Mention & Top-2 Similar Instances & Prediction \\
        \midrule
        \multirow{2}*{\parbox{1.5cm}{feel uncomfortable}} & \multirow{2}*{\parbox{3.3cm}{
        (felt uncomfortable, uneasy) (uncomfortable, uneasy)
        }} & ($-$) discomfort {\color{red} $\times$} \\
        & & ($+$) uneasy {\color{blue} $\checkmark$} \\
        \midrule
        \multirow{2}*{\parbox{1.5cm}{blurred vision}} & \multirow{2}*{\parbox{3.3cm}{
        (blurry vision, misty vision) (blurry, blurring of visual image)
        }} & ($-$) double vision {\color{red} $\times$} \\
        & & ($+$) misty vision {\color{blue} $\checkmark$} \\
        \bottomrule
    \end{tabular}
    \caption{Two test cases from COMETA with predictions made without ($-$) or with ($+$) $k$NN Module.}
    \label{tab:case-study}
\end{table}

Apart from the long-tailed case study, we provide case analysis between BioELQA and Prompt-BioEL.
Table~\ref{tab:case-study-comparison-1} lists four cases of BioELQA predicting correctly and Prompt-BioEL predicting incorrectly. BioELQA performs better in handling long-tailed entities or entities with morphological similarities because it can retrieve answers from similar mentions in the training set as references.
In contrast, Prompt-BioEL performs better in handling ambiguous mentions because it incorporates additional context to disambiguate, while BioELQA does not consider contextual information, as shown in Table~\ref{tab:case-study-comparison-2} where Prompt-BioEL predicts correctly and BioELQA predicts incorrectly.

\begin{table}[h]
    \centering
    \scriptsize
    \renewcommand\arraystretch{1.0}
    \begin{tabular}{@{}l|l|p{2.4cm}@{}}
        \toprule
        Mention & Retrieved Instances & Prediction \\
        \midrule
        \multirow{3}*{\parbox{1.08cm}{seizure medication}} & \multirow{3}*{\parbox{3.3cm}{
        (seizure meds, seizure management), (migraine treatments, migraine prophylaxis)
        }} & ($Q$) seizure management \\
        & & ($P$) seizure finding \\
        \midrule
        \multirow{2}*{\parbox{1.08cm}{sketchy}} & \multirow{2}*{\parbox{3.3cm}{
        (sketchy, incomplete), (blurry, blurring of visual image)
        }} & ($Q$) incomplete \\
        & & ($P$) brief \\
        \midrule
        \multirow{3}*{\parbox{1.08cm}{rhodiola}} & \multirow{3}*{\parbox{3.3cm}{
        (rhodiola, family crassulaceae), (broccoli, brassica oleracea)
        }} & ($Q$) family crassulaceae \\
        & & ($P$) ralstonia \\
        \midrule
        \multirow{2}*{\parbox{1.08cm}{lose weight}} & \multirow{2}*{\parbox{3.3cm}{
        (losing weight, weight loss)
        }} & ($Q$) weight loss \\
        & & ($P$) weight decreasing \\
        \bottomrule
    \end{tabular}
    \caption{Four test cases on COMETA dataset where BioELQA ($Q$) made correct predictions and Prompt-BioEL ($P$) did not.}
    \label{tab:case-study-comparison-1}
\end{table}

\begin{table}[h]
    \centering
    \scriptsize
    \renewcommand\arraystretch{1.0}
    \begin{tabular}{@{}l|l|p{2.3cm}@{}}
        \toprule
        Mention & Context & Prediction \\
        \midrule
        \multirow{3}*{\parbox{1.2cm}{suppositories}} & \multirow{3}*{\parbox{3.3cm}{
        for some personal reasons i can't use the [E1] suppositories [/E1]
        }} & ($Q$) suppository \\
        & & ($P$) suppository physical object \\
        \midrule
        \multirow{3}*{\parbox{1.2cm}{salt}} & \multirow{3}*{\parbox{3.3cm}{
        i guess maybe some [E1] salt [/E1] residue gets on the insertion site
        }} & ($Q$) salt water \\
        & & ($P$) sodium chloride \\
        & & \\
        \midrule
        \multirow{3}*{\parbox{1.2cm}{sublingual}} & \multirow{3}*{\parbox{3.3cm}{maxalt comes in a [E1] sublingual [/E1] under the tongue disolve version
        }} & ($Q$) sublingual \\
        & & ($P$) sublingual intended site \\
        \midrule
        \multirow{5}*{\parbox{1.2cm}{gum}} & \multirow{5}*{\parbox{3.3cm}{
        can't really see it but there's a dark line underneath my front teeth what seems like around my [E1] gum [/E1] shown above
        }} & ($Q$) gum \\
        & & ($P$) gingival structure \\
        & & \\
        & & \\
        & & \\
        \bottomrule
    \end{tabular}
    \caption{Four test cases on COMETA dataset where Prompt-BioEL ($P$) made correct predictions and BioELQA ($Q$) did not.}
    \label{tab:case-study-comparison-2}
\end{table}

\subsection{Impacts of hyper-parameters}
\begin{figure}[!h]
    \centering
    \includegraphics[width=0.9\linewidth]{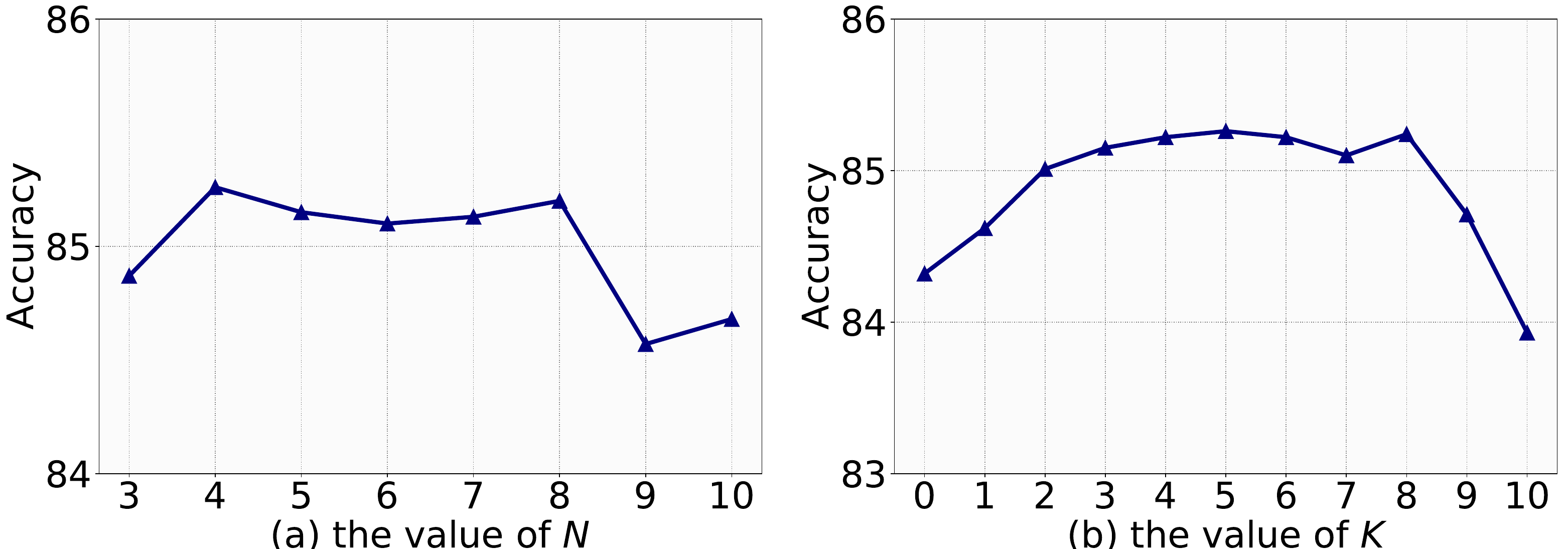}
    \caption{Impact of different hyper-parameters on the COMETA dataset.}
    \label{fig:analysis}
\end{figure}

In Figure~\ref{fig:analysis}(a), we observe that the accuracy initially improves and then declines as $N$ increases. A higher $N$ indicates a higher recall rate but also introduces more wrong options, leading to decreased performance.
Similarly, in Figure~\ref{fig:analysis}(b), with the increase of $K$, accuracy increases first and decreases afterward. This suggests that allowing too many retrieved instances can potentially confuse the model and negatively impact predictions.

\section{Conclusion}
\label{sec:con}


We propose BioELQA, a new MCQA-based method that introduces a multiple-choice prompt that allows for more comprehensive understanding of the interactions between different entities and a given mention.
In addition, a $k$NN module that recaptures related training data as a side-by-side reminder can explicitly provide essential information to enhance performance.
We experimentally validate the state-of-the-art performance of our method in several public datasets.
For future work, we plan to explore introducing contextual information to BioELQA to address entity ambiguation issues.

\newpage
\balance
\section{Bibliographical References}\label{sec:reference}

\bibliographystyle{lrec-coling2024-natbib}
\bibliography{refs}

\begin{thebibliography}{36}
\expandafter\ifx\csname natexlab\endcsname\relax\def\natexlab#1{#1}\fi

\bibitem[{Angell et~al.(2021)Angell, Monath, Mohan, Yadav, and
  McCallum}]{angell2021clustering}
Rico Angell, Nicholas Monath, Sunil Mohan, Nishant Yadav, and Andrew McCallum.
  2021.
\newblock {Clustering-based Inference for Biomedical Entity Linking}.
\newblock In \emph{Proceedings of the 2021 Conference of the North American
  Chapter of the Association for Computational Linguistics: Human Language
  Technologies}, pages 2598--2608.

\bibitem[{Anil et~al.(2023)Anil, Dai, Firat, Johnson, Lepikhin, Passos,
  Shakeri, Taropa, Bailey, Chen et~al.}]{anil2023palm}
Rohan Anil, Andrew~M Dai, Orhan Firat, Melvin Johnson, Dmitry Lepikhin,
  Alexandre Passos, Siamak Shakeri, Emanuel Taropa, Paige Bailey, Zhifeng Chen,
  et~al. 2023.
\newblock {PaLM} 2 technical report.
\newblock \emph{arXiv preprint arXiv:2305.10403}.

\bibitem[{Basaldella et~al.(2020)Basaldella, Liu, Shareghi, and
  Collier}]{basaldella2020cometa}
Marco Basaldella, Fangyu Liu, Ehsan Shareghi, and Nigel Collier. 2020.
\newblock {COMETA}: {A} {C}orpus for {M}edical {E}ntity {L}inking in the
  {S}ocial {M}edia.
\newblock In \emph{Proceedings of the 2020 Conference on Empirical Methods in
  Natural Language Processing (EMNLP)}, pages 3122--3137.

\bibitem[{Bodenreider(2004)}]{bodenreider2004unified}
Olivier Bodenreider. 2004.
\newblock The {U}nified {M}edical {L}anguage {S}ystem ({UMLS}): integrating
  biomedical terminology.
\newblock \emph{Nucleic Acids Research}, 32.

\bibitem[{Brown et~al.(2020)Brown, Mann, Ryder, Subbiah, Kaplan, Dhariwal,
  Neelakantan, Shyam, Sastry, Askell et~al.}]{brown2020language}
Tom Brown, Benjamin Mann, Nick Ryder, Melanie Subbiah, Jared~D Kaplan, Prafulla
  Dhariwal, Arvind Neelakantan, Pranav Shyam, Girish Sastry, Amanda Askell,
  et~al. 2020.
\newblock {Language Models are Few-shot Learners}.
\newblock \emph{Advances in {N}eural {I}nformation {P}rocessing {S}ystems},
  33:1877--1901.

\bibitem[{Do{\u{g}}an et~al.(2014)Do{\u{g}}an, Leaman, and Lu}]{dougan2014ncbi}
Rezarta~Islamaj Do{\u{g}}an, Robert Leaman, and Zhiyong Lu. 2014.
\newblock {NCBI} {D}isease {C}orpus: {A} {R}esource for {D}isease {N}ame
  {R}ecognition and {C}oncept {N}ormalization.
\newblock \emph{Journal of Biomedical Informatics}, 47.

\bibitem[{Gao et~al.(2021)Gao, Yao, and Chen}]{gao2021simcse}
Tianyu Gao, Xingcheng Yao, and Danqi Chen. 2021.
\newblock {S}im{CSE}: {S}imple {C}ontrastive {L}earning of {S}entence
  {E}mbeddings.
\newblock In \emph{Proceedings of the 2021 Conference on Empirical Methods in
  Natural Language Processing}.

\bibitem[{Jahan et~al.(2023)Jahan, Laskar, Peng, and
  Huang}]{jahan2023comprehensive}
Israt Jahan, Md~Tahmid~Rahman Laskar, Chun Peng, and Jimmy Huang. 2023.
\newblock A {C}omprehensive {E}valuation of {L}arge {L}anguage {M}odels on
  {B}enchmark {B}iomedical {T}ext {P}rocessing {T}asks.
\newblock \emph{arXiv preprint arXiv:2310.04270}.

\bibitem[{Johnson et~al.(2019)Johnson, Douze, and
  J{\'e}gou}]{johnson2019billion}
Jeff Johnson, Matthijs Douze, and Herv{\'e} J{\'e}gou. 2019.
\newblock Billion-{S}cale {S}imilarity {S}earch with {GPU}s.
\newblock \emph{IEEE Transactions on Big Data}, 7(3).

\bibitem[{Khandelwal et~al.(2020)Khandelwal, Fan, Jurafsky, Zettlemoyer, and
  Lewis}]{khandelwal2020nearest}
Urvashi Khandelwal, Angela Fan, Dan Jurafsky, Luke Zettlemoyer, and Mike Lewis.
  2020.
\newblock Nearest {N}eighbor {M}achine {T}ranslation.
\newblock In \emph{International Conference on Learning Representations}.

\bibitem[{Kim and Ganapathi(2021)}]{kim2021read}
Byung-Hak Kim and Varun Ganapathi. 2021.
\newblock {Read, Attend, and Code: Pushing the Limits of Medical Codes
  Prediction from Clinical Notes by Machines}.
\newblock In \emph{Machine Learning for Healthcare Conference}, pages 196--208.
  PMLR.

\bibitem[{Lai et~al.(2021)Lai, Ji, and Zhai}]{lai2021bert}
Tuan Lai, Heng Ji, and ChengXiang Zhai. 2021.
\newblock {BERT} might be {O}verkill: A {T}iny but {E}ffective {B}iomedical
  {E}ntity {L}inker based on {R}esidual {C}onvolutional {N}eural {N}etworks.
\newblock In \emph{Findings of the Association for Computational Linguistics:
  EMNLP}.

\bibitem[{Lewis et~al.(2020)Lewis, Liu, Goyal, Ghazvininejad, Mohamed, Levy,
  Stoyanov, and Zettlemoyer}]{lewis2020bart}
Mike Lewis, Yinhan Liu, Naman Goyal, Marjan Ghazvininejad, Abdelrahman Mohamed,
  Omer Levy, Veselin Stoyanov, and Luke Zettlemoyer. 2020.
\newblock {BART}: {D}enoising {S}equence-to-{S}equence {P}re-training for
  {N}atural {L}anguage {G}eneration, {T}ranslation, and {C}omprehension.
\newblock In \emph{Proceedings of the 58th Annual Meeting of the Association
  for Computational Linguistics}.

\bibitem[{Li et~al.(2016)Li, Sun, Johnson, Sciaky, Wei, Leaman, Davis,
  Mattingly, Wiegers, and Lu}]{li2016biocreative}
Jiao Li, Yueping Sun, Robin~J Johnson, Daniela Sciaky, Chih-Hsuan Wei, Robert
  Leaman, Allan~Peter Davis, Carolyn~J Mattingly, Thomas~C Wiegers, and Zhiyong
  Lu. 2016.
\newblock Bio{C}reative {V} {CDR} task corpus: a resource for chemical disease
  relation extraction.
\newblock \emph{Database}.

\bibitem[{Li et~al.(2023)Li, Qiu, Zhao, Zhang, Xing, and Wu}]{li2023drug}
Xinhang Li, Zhaopeng Qiu, Xiangyu Zhao, Yong Zhang, Chunxiao Xing, and Xian Wu.
  2023.
\newblock {REST: Drug-Drug Interaction Prediction via Reinforced
  Student-Teacher Curriculum Learning}.
\newblock In \emph{Proceedings of the 32nd ACM International Conference on
  Information and Knowledge Management}, CIKM '23, page 1278–1287.
  Association for Computing Machinery.

\bibitem[{Lin et~al.(2022)Lin, Zhang, Wang, Shi, Wu, and Zheng}]{lin2022multi}
Zhenxi Lin, Ziheng Zhang, Meng Wang, Yinghui Shi, Xian Wu, and Yefeng Zheng.
  2022.
\newblock {Multi-modal Contrastive Representation Learning for Entity
  Alignment}.
\newblock In \emph{Proceedings of the 29th International Conference on
  Computational Linguistics}, pages 2572--2584.

\bibitem[{Lin et~al.(2024)Lin, Zhang, Wu, and Zheng}]{lin2023improving}
Zhenxi Lin, Ziheng Zhang, Xian Wu, and Yefeng Zheng. 2024.
\newblock \href {https://doi.org/10.1109/ICASSP48485.2024.10448513} {Improving
  biomedical entity linking with retrieval-enhanced learning}.
\newblock In \emph{ICASSP 2024 - 2024 IEEE International Conference on
  Acoustics, Speech and Signal Processing (ICASSP)}, pages 11461--11465.

\bibitem[{Liu et~al.(2021)Liu, Shareghi, Meng, Basaldella, and
  Collier}]{liu2021self}
Fangyu Liu, Ehsan Shareghi, Zaiqiao Meng, Marco Basaldella, and Nigel Collier.
  2021.
\newblock Self-{A}lignment {P}retraining for {B}iomedical {E}ntity
  {R}epresentations.
\newblock In \emph{Proceedings of the 2021 Conference of the North American
  Chapter of the Association for Computational Linguistics: Human Language
  Technologies}.

\bibitem[{Pezeshkpour and Hruschka(2023)}]{pezeshkpour2023large}
Pouya Pezeshkpour and Estevam Hruschka. 2023.
\newblock {Large Language Models Sensitivity to The Order of Options in
  Multiple-Choice Questions}.
\newblock \emph{arXiv preprint arXiv:2308.11483}.

\bibitem[{Raffel et~al.(2020)Raffel, Shazeer, Roberts, Lee, Narang, Matena,
  Zhou, Li, and Liu}]{raffel2020exploring}
Colin Raffel, Noam Shazeer, Adam Roberts, Katherine Lee, Sharan Narang, Michael
  Matena, Yanqi Zhou, Wei Li, and Peter~J Liu. 2020.
\newblock {Exploring the Limits of Transfer Learning with a Unified
  Text-to-Text Transformer}.
\newblock \emph{The Journal of Machine Learning Research}, 21(1):5485--5551.

\bibitem[{Robinson et~al.(2022)Robinson, Rytting, and
  Wingate}]{robinson2022leveraging}
Joshua Robinson, Christopher~Michael Rytting, and David Wingate. 2022.
\newblock {Leveraging Large Language Models for Multiple Choice Question
  Answering}.
\newblock \emph{arXiv preprint arXiv:2210.12353}.

\bibitem[{Shi et~al.(2022)Shi, Zhao, Wang, Chen, Zhang, Zheng, and
  Che}]{shi2022understand}
Xiaoming Shi, Sendong Zhao, Yuxuan Wang, Xi~Chen, Ziheng Zhang, Yefeng Zheng,
  and Wanxiang Che. 2022.
\newblock {Understanding Patient Query With Weak Supervision From Doctor
  Response}.
\newblock \emph{IEEE Journal of Biomedical and Health Informatics},
  26(6):2770--2777.

\bibitem[{Sung et~al.(2020)Sung, Jeon, Lee, and Kang}]{sung2020biomedical}
Mujeen Sung, Hwisang Jeon, Jinhyuk Lee, and Jaewoo Kang. 2020.
\newblock Biomedical {E}ntity {R}epresentations with {S}ynonym
  {M}arginalization.
\newblock In \emph{Proceedings of the 58th Annual Meeting of the Association
  for Computational Linguistics}, pages 3641--3650.

\bibitem[{Sutskever et~al.(2014)Sutskever, Vinyals, and
  Le}]{sutskever2014sequence}
Ilya Sutskever, Oriol Vinyals, and Quoc~V Le. 2014.
\newblock Sequence to {S}equence {L}earning with {N}eural {N}etworks.
\newblock \emph{Advances in {N}eural {I}nformation {P}rocessing {S}ystems}, 27.

\bibitem[{Touvron et~al.(2023)Touvron, Martin, Stone, Albert, Almahairi,
  Babaei, Bashlykov, Batra, Bhargava, Bhosale et~al.}]{touvron2023llama}
Hugo Touvron, Louis Martin, Kevin Stone, Peter Albert, Amjad Almahairi, Yasmine
  Babaei, Nikolay Bashlykov, Soumya Batra, Prajjwal Bhargava, Shruti Bhosale,
  et~al. 2023.
\newblock Llama 2: {O}pen {F}oundation and {F}ine-tuned {C}hat {M}odels.
\newblock \emph{arXiv preprint arXiv:2307.09288}.

\bibitem[{Varma et~al.(2021)Varma, Orr, Wu, Leszczynski, Ling, and
  R{\'e}}]{varma2021cross}
Maya Varma, Laurel Orr, Sen Wu, Megan Leszczynski, Xiao Ling, and Christopher
  R{\'e}. 2021.
\newblock Cross-{D}omain {D}ata {I}ntegration for {N}amed {E}ntity
  {D}isambiguation in {B}iomedical {T}ext.
\newblock In \emph{Findings of the Association for Computational Linguistics:
  EMNLP 2021}, pages 4566--4575.

\bibitem[{Wu et~al.(2020)Wu, Petroni, Josifoski, Riedel, and
  Zettlemoyer}]{wu2020scalable}
Ledell Wu, Fabio Petroni, Martin Josifoski, Sebastian Riedel, and Luke
  Zettlemoyer. 2020.
\newblock Scalable {Z}ero-shot {E}ntity {L}inking with {D}ense {E}ntity
  {R}etrieval.
\newblock In \emph{Proceedings of the 2020 Conference on Empirical Methods in
  Natural Language Processing}, pages 6397--6407.

\bibitem[{Wu et~al.(2023)Wu, Bai, Guo, Liu, Li, and Yang}]{wu2023modeling}
Taiqiang Wu, Xingyu Bai, Weigang Guo, Weijie Liu, Siheng Li, and Yujiu Yang.
  2023.
\newblock {Modeling Fine-grained Information via Knowledge-aware Hierarchical
  Graph for Zero-shot Entity Retrieval}.
\newblock In \emph{Proceedings of the Sixteenth ACM International Conference on
  Web Search and Data Mining}, pages 1021--1029.

\bibitem[{Xiang et~al.(2021)Xiang, Zhang, Chen, Chen, Lin, and
  Zheng}]{xiang2021ontoea}
Yuejia Xiang, Ziheng Zhang, Jiaoyan Chen, Xi~Chen, Zhenxi Lin, and Yefeng
  Zheng. 2021.
\newblock {OntoEA: Ontology-guided Entity Alignment via Joint Knowledge Graph
  Embedding}.
\newblock In \emph{Findings of the Association for Computational Linguistics:
  ACL-IJCNLP 2021}, pages 1117--1128.

\bibitem[{Xu et~al.(2023)Xu, Chen, and Hu}]{xu2023improving}
Zhenran Xu, Yulin Chen, and Baotian Hu. 2023.
\newblock Improving {B}iomedical {E}ntity {L}inking with {C}ross-{E}ntity
  {I}nteraction.
\newblock In \emph{Proceedings of the AAAI Conference on Artificial
  Intelligence}, volume~37.

\bibitem[{Yan et~al.(2020)Yan, Wang, Xiang, Zhou, and Zong}]{yan2020knowledge}
Jinghui Yan, Yining Wang, Lu~Xiang, Yu~Zhou, and Chengqing Zong. 2020.
\newblock {A Knowledge-driven Generative Model for Multi-implication Chinese
  Medical Procedure Entity Normalization}.
\newblock In \emph{Proceedings of the 2020 Conference on Empirical Methods in
  Natural Language Processing}, pages 1490--1499.

\bibitem[{Yuan and Yu(2021)}]{yuan2021efficient}
Hongyi Yuan and Sheng Yu. 2021.
\newblock Efficient {S}ymptom {I}nquiring and {D}iagnosis via {A}daptive
  {A}lignment of {R}einforcement {L}earning and {C}lassification.
\newblock \emph{arXiv preprint arXiv:2112.00733}.

\bibitem[{Yuan et~al.(2022{\natexlab{a}})Yuan, Yuan, Gan, Zhang, Xie, and
  Yu}]{yuan2022biobart}
Hongyi Yuan, Zheng Yuan, Ruyi Gan, Jiaxing Zhang, Yutao Xie, and Sheng Yu.
  2022{\natexlab{a}}.
\newblock {B}io{BART}: {P}retraining and {E}valuation of a {B}iomedical
  {G}enerative {L}anguage {M}odel.
\newblock \emph{arXiv preprint arXiv:2204.03905}.

\bibitem[{Yuan et~al.(2022{\natexlab{b}})Yuan, Yuan, and
  Yu}]{yuan2022generative}
Hongyi Yuan, Zheng Yuan, and Sheng Yu. 2022{\natexlab{b}}.
\newblock Generative {B}iomedical {E}ntity {L}inking via {K}nowledge
  {B}ase-{G}uided {P}re-training and {S}ynonyms-{A}ware {F}ine-tuning.
\newblock In \emph{Proceedings of the 2022 Conference of the North American
  Chapter of the Association for Computational Linguistics: Human Language
  Technologies}.

\bibitem[{Zhang et~al.(2023)Zhang, Yao, Yue, Wu, Zhang, Lin, and
  Zheng}]{zhang2023emerging}
Yongqi Zhang, Quanming Yao, Ling Yue, Xian Wu, Ziheng Zhang, Zhenxi Lin, and
  Yefeng Zheng. 2023.
\newblock Emerging {D}rug {I}nteraction {P}rediction {E}nabled by {F}low-based
  {G}raph {N}eural {N}etwork with {B}iomedical {N}etwork.
\newblock \emph{Nature Computational Science}, 3(12):1023--1033.

\bibitem[{Zhu et~al.(2021)Zhu, Qin, Chen, Hu, and Xiang}]{zhu2021enhancing}
Tiantian Zhu, Yang Qin, Qingcai Chen, Baotian Hu, and Yang Xiang. 2021.
\newblock Enhancing {E}ntity {R}epresentations with {P}rompt {L}earning for
  {B}iomedical {E}ntity {L}inking.
\newblock In \emph{Proceedings of the Thirty-First International Joint
  Conference on Artificial Intelligence}.

\end{thebibliography}


\end{document}